\documentclass[conference]{IEEEtran}

\usepackage{cite}
\usepackage{amsmath,amssymb,amsfonts}
\usepackage{amsthm}
\usepackage[ruled]{algorithm2e}
\usepackage{graphicx}
\usepackage{textcomp}
\usepackage{xcolor}
\usepackage{subcaption}
\usepackage{multirow}
\usepackage{booktabs}
\usepackage{makecell}
\newcommand{\argmax}{\mathop{\mathrm{argmax}}}
\begin{document}

\title{Exposing Surveillance Detection Routes via Reinforcement Learning, Attack Graphs, and Cyber Terrain}

\author{Lanxiao Huang$^{a}$$^{*}$,
        Tyler Cody$^{a}$,
        Christopher Redino$^{b}$,
        Abdul Rahman$^{b}$,\\
        Akshay Kakkar$^{b}$,
        Deepak Kushwaha$^{b}$,
        Cheng Wang$^{b}$,
        Ryan Clark$^{b}$,\\
        Daniel Radke$^{b}$,
        Peter Beling$^{a}$,
        Edward Bowen$^{b}$\\
        \small $^{a}$National Security Institute, Virginia Polytechnic University \\
        \small $^{b}$Deloitte \& Touche LLP \\
        \small $^{*}$Corresponding author: Lanxiao Huang: hlanxiao@vt.edu \\
}
\maketitle

\begin{abstract}

Reinforcement learning (RL) operating on attack graphs leveraging cyber terrain principles are used to develop reward and state associated with determination of surveillance detection routes (SDR). This work extends previous efforts on developing RL methods for path analysis within enterprise networks. This work focuses on building SDR where the routes focus on exploring the network services while trying to evade risk. RL is utilized to support the development of these routes by building a reward mechanism that would help in realization of these paths. The RL algorithm is modified to have a novel warm-up phase which decides in the initial exploration which areas of the network are safe to explore based on the rewards and penalty scale factor.

\end{abstract}

\begin{IEEEkeywords}
attack graphs, reinforcement learning, surveillance detection routes, SDR, cyber terrain
\end{IEEEkeywords}

\section{Introduction}


Reconnaissance (also called recon) in MITRE's Adversarial Tactics, Techniques, and Common Knowledge (ATT\&CK\textregistered) framework is described as "techniques that involve adversaries actively or passively gathering information that can be used to support targeting \cite{mitre-attack}." As reconnaissance activities usually precede an exploitation campaign, detection of these efforts benefit cyber defenders by identifying potential targets (e.g., crown jewels) of interest. In this respect, adversarial recon activities strive to maximize visibility of targets while minimizing opportunities of being detected. Critical to this are the identification of paths, termed SDR, traversed by adversaries to gather critical data about targets (e.g., ports, protocols, applications, services).

From a cyber protection standpoint, recon activities disguise serious hostile intent but may be quite difficult to quantitatively differentiate from normal behavior. Malicious intent is quite difficult to observe, as it may be efficiently designed to hide in the background of normal traffic. Detecting this type of recon is situational in nature and requires meticulous analyses of huge volumes of collected data. Other domains apart from cyber are often challenged in a similar manner, where evaluation of these SDR require data analysis to differentiate abnormal traffic from events that are suspicious in proximity to roads and crossings \cite{route-surveillance}.


Modern efforts to detect and respond to adversarial network reconnaissance are a complicated blend of automated and human processes. Automated collection systems are installed on network devices and endpoints to passively and actively monitor the network communications, analyze the flow, and aggregate the data for the Security Information Event Management (SIEM) and/or Security Orchestration, Automation and Response (SOAR) systems for analysis.  These network tools assist the human component of detection by providing automated security reports, incident alerts, and executing network protection protocols with a single click from the security operations center (SOC) analyst. The effectiveness of these systems relies on the data collected, the knowledge of current threat behavior, and the human analyst’s ability to understand the threat. Naturally, such approaches have blindspots. 

Combining the current security information (network topology and configuration) with machine learning (ML) analysis allows highlighting weak points missed by automated systems that an attacker may focus on during initial recon. Network traffic behavior analysis, no matter how advanced, relies on active network traffic and does not preempt network/host/protocol mis-configurations. This paper contributes a deep RL approach to generating SDR in the form of attack graphs from network models consisting of network topology and configuration, thereby extending the suite of automated tools and systems available for the cyber defense.

In particular, an Markov decision process (MDP) formulation and a new algorithm, SDR-RL, is proposed that uses a warm-up phase and penalty scaling to control the asymmetry between the number of host services scanned and the number of defensive terrain encountered. This emulates the asymmetry sought by human operators when conducting network reconnaissance generally and SDR in particular. Additionally, this paper extends the double agent architecture (DAA) of Nguyen et al. \cite{nguyen2021proposal}, which originally used standard deep Q-learning (DQN), with actor-critic (A2C) and proximal-policy optimization (PPO) algorithms\cite{schulman2017proximal}.

This paper is structured as follows. First, background on RL and penetration testing is given. Then, the methods used to expose SDR using RL, attack graphs, and cyber terrain are presented. Next, experimental design for testing the presented methodology is given, including details regarding RL implementation and training. Results are presented, and, before concluding, an in depth discussion is given in terms of cyber-specific outcomes. 


\section{RL and Penetration Testing}

\subsection{Reinforcement Learning Preliminaries}
Reinforcement learning (RL) problems involve an agent, interacting with an environment, and transiting from one state to another until it reaches the goal state \cite{sutton2018reinforcement}. The agent receives rewards for taking actions with the overall goal to achieve maximum cumulative rewards. Environments are normally modeled as MDPs, which can be defined by a tuple $\{S,A,R,P,\gamma\}$ where S denotes the set of possible states and A denotes the set of possible actions, R represents the distribution of reward given any state-action pair, P represents transition probability and $\gamma$ is discount factor. The goal of the agent is to learn an optimal policy $\pi$ that maps states to actions. As so, the RL algorithm is to learn an optimal policy to select actions and maximize the expected cumulative discounted reward:
\begin{equation}
\begin{split}
    \pi^* =\argmax_{\pi}\mathbb{E}[\sum_{t>0} \gamma^t r_t |\pi]\\
    with\;s_0\sim p(s_0),a_t\sim \pi(\cdot|s_0),s_{t+1}\sim p(\cdot|s_t,a_t)
\end{split}
\end{equation}

In (deep) Q-learning, the optimal policy can be defined by two terms, the value function and Q-value function. The value function shows how good the state is and it is defined as the expected cumulative reward from following the policy from the state
\begin{equation}
    V^\pi (s)=\mathbb{E}[\sum_{t>0} \gamma^t r_t |\pi]
\end{equation}

The Q-value function, on the other side, is defined as the expected cumulative reward given both parts of the state-action pair
\begin{equation}
    Q^\pi (s,a)=\mathbb{E}[\sum_{t>0} \gamma^t r_t |s_0=s,a_0=a,\pi]
\end{equation}

It satisfies the Bellman equation
\begin{equation}
    Q^* (s,a)=\mathbb{E}_s\sim\epsilon [r+\gamma\max_{a'}Q^* (s',a') |s,a]
\end{equation}

Thus, the optimal policy $\pi^*$ corresponds to taking the best action in any state as specified by $Q^*$.

For a more complex problem where the state or action space is large enough so that computing every state-action pair is infeasible, neural networks become a powerful function estimator in deep Q-learning (DQL) \cite{mnih2013playing, mnih2015human}. The optimal Q-values are estimated by a neural network parameterized by $\theta$:
\begin{equation}
    Q(s,a;\theta) \approx Q^* (s,a)
\end{equation}
where $y_i=\mathbb{E}_{s'\sim\epsilon} [r+\gamma\max_{a'} Q^* (s',a';\theta_{i-1}) |s,a]$

During the backward pass, the gradients can be calculated as

\begin{align}
    \nabla_{\theta_i}L_i(\theta_i)=& \nonumber\\
    &\mathbb{E}_{s,a\sim p(\cdot);s\sim\epsilon}[
    (r+\gamma\max_{a'} Q^* (s',a';\theta_{i-1}) \nonumber\\
    &-Q(s,a;\theta_i))\nabla_{\theta_i}Q(s,a;\theta_i)]
\end{align}

\subsection{The Penetration Testing Environment}

Though RL has been pursued as a tool for penetration testing recently, the approaches to model the network vary significantly. Alternative methods to modeling penetration testing including hypothesis generation, ontology-based, attack trees and attack graphs. Hypothesis generation model \cite{weissman1995penetration} is an organizational network presentation for cyber defense while ontology-based models \cite{chu2020ontology} focus more on the semantics of penetration testing. However, neither of them contains any structural information about the network itself. While both attack trees \cite{schneier1999attack} and attack graphs do provide structural representation of the network, attack graphs are more generative and attack trees are special cases of attack graphs. As a result, attack graph are the most recognized method for the modeling of penetration testing environment \cite{cody2022layered}.

\subsection{Related Work}

The use of deep RL for attack graphs has seen recent development. Other than Ghanem and Chen \cite{ghanem2020reinforcement}, the authors in the RL for attack graphs literature use fully observable MDPs to model networks. Many authors use the Common Vulnerability Scoring System (CVSS) to furnish their MDPs (CVSS-MDPs). Yousefi et al. provide the earliest work doing so in deep RL for penetration testing \cite{yousefi2018reinforcement}. Hu et al. extend the use of the CVSS by proposing to use exploitability scores weight rewards \cite{hu2020automated}. Gangupantulu et al. \cite{gangupantulu2021using, gangupantulu2021crown} and Cody et al. \cite{cody2022discovering} explicitly extend the methods of Hu et al. with concepts of terrain. Gangupantulu et al. advocate defining models of terrain in terms of the rewards and transition probabilities of MDPs, first in the case of firewalls as obstacles \cite{gangupantulu2021using}, then in the case of lateral pivots nearby key terrain \cite{gangupantulu2021crown}. Cody et al. apply these concepts to exfiltration \cite{cody2022discovering}. Other authors either handcraft the MDP or do not remark on how its components are estimated.

Many authors apply generic deep Q learning (DQN) \cite{mnih2013playing, mnih2015human} to solve point-to-point network traversal tasks \cite{yousefi2018reinforcement, schwartz2019autonomous, chowdhary2020autonomous, hu2020automated, gangupantulu2021using}. Typically the terminal state is unknown and solutions take the form of individual paths. Others develop domain-specific modifications for deep RL including the double agent architecture \cite{nguyen2021proposal}, a hierarchical action decomposition approach \cite{tran2021deep}, and various improvements to DQN termed NDSPI-DQN \cite{zhou2021autonomous}. Another line of research focuses on developing more specific penetration testing tasks. A number of authors define more specific tasks by reward engineering and other modifications to the MDP including formulations of capture the flag \cite{zennaro2020modeling}, crown jewel analysis \cite{gangupantulu2021crown}, and discovering exfiltration paths \cite{cody2022discovering}. This paper extends this line of research with a methodology for exposing SDR.

\section{Methods}

The following subsections describe the presented methods for adding service-based risk penalties as defensive terrain in CVSS-MDPs and the algorithm for discovering surveillance paths in a network.

\subsection{Defensive Terrain in CVSS-MDPs}

Gangupantulu et al. \cite{gangupantulu2021using} proposed that cyber terrain can be modeled into CVSS-MDPs by adding transition probabilities for traversing firewalls and negative rewards for different protocols. Cody et al. \cite{cody2022discovering} later modeled the services-based defensive terrain in CVSS-MDPs based on the assumption that the attackers can infer the presence of defenses terrain based on the services running on a host. We adopt their methods and classify the services into four categories: authentication, data, security and common. Our reward hierarchy assigns -6 for authentication, -4 for data and -2 for security and common. Moreover, the type of actions has an effect on the rewards. n+1 (-1, -3 and -5) is assigned for scanning action while n (-2, -4 and -6) is assigned for exploiting action. 

\subsection{Discovering Surveillance Paths with RL}

In our surveillance detection routes (SDR) algorithm (Algorithm 1), we have a target node of interest which we want to explore all the services running on it. The goal of SDR is to gain the service information of the target node along with maximizing service information discovery along other areas of the network while being cautious and not triggering any defensive terrain area. We give a discovery reward of 100 to the target node when all its service have been discovered. To encourage the service discovery the agent receives a +1 reward for each service discovered on a node. The algorithm is divided into two phases - a warm-up phase and a training phase.

\subsection{Warmup Phase}
In the warmup phase, we want to update our goal for the RL agent to not only gain information of just the target node but also other nodes from where we receive positive reward indicating that the defense terrain allows us to access this area. The following steps are taken in the warmup phase:
\begin{itemize}
    \item We define a certain number of episodes for warmup phase in the training configuration.
    \item In the warmup episodes the RL agent does not learn (no weight updates) but only monitors positive reward during an episode.
    \item If the RL agent receives a positive reward from a scanning action, then that node is added to the goal along with the target node and its reward for compromise is set to 100.
    \item The algorithm allows only one node (which gives out the maximum positive reward) from each subnet to be part of the goal as gaining control of one node in a subnet is enough to gain service information of the entire subnet.
\end{itemize}

At the end of the warmup phase, we have these dynamic nodes along with the target node of which we must gain the service information to reach the goal. The number of dynamic nodes being a part of the goal decreases as the scale value of defense terrain increases. Therefore, we limit the agent’s exploration capability as we increase the scale value of defense terrain.

\subsection{Training Phase}
In the training phase, the agent interacts with network in an episodic fashion to learn which is the best possible path to gain service information of all the target nodes identified during the warmup phase.

\subsection{Scalability}
Conventionally, DQN is the most basic algorithm used for modeling RL on penetration testing. Nguyen at al. proposed a method that utilizes two A2C agents: one called the structural agent that learns the structural information about subnets, hosts and firewalls as well as the connections between them. The other called the exploiting agent selects actions to take and the target of them. Their method solves the scalability problem of DQN to some extent and has a better capacity for large networks. We improve upon this method by applying proximal policy optimization (PPO) instead of A2C for both of the agents. PPO is an advanced RL algorithm known for its convergence speed, stability and sample efficiency. It optimizes the following clipped surrogate objective function to prevent performance collapse caused by a large policy update:
\begin{align}
    \mathcal{L}(\theta) = \mathbb{E} \Big[ \min\big(\rho_t(\theta) A_t, \mathrm{clip}\big( \rho_t(\theta), 1-\epsilon, 1+ \epsilon \big) A_t\big)\Big], \label{eq:ppo_obj}
\end{align}
where $\rho_t(\theta) = \pi_\theta(a_t|s_t) / \pi_{\theta_{\mathrm{old}}} (a_t|s_t)$ is the probability ratio of the new policy over the old policy. The advantage function $A_t$  is often estimated using the generalized advantage estimation \cite{schulman2015high}, truncated after $T$ steps:
\begin{align}
    & A_t \approx \delta_t + (\gamma\lambda) \delta_{t+1}+\cdots + (\gamma\lambda)^{T-t+1}\delta_{T-1},\\
    & \mathrm{where\;} \delta_t = r_t + \gamma V(s_{t+1}) - V(s_t).
\end{align}

\begin{algorithm}[ht]
    \SetKwInOut{Input}{input}
    \SetKwInOut{Output}{output}
    \caption{SDR via RL (SDR-RL)}\label{algorithm}

    \Input{MDP $\mathcal{M}$, initial node $i$, set of target nodes $J$, number of warmup episode $N_w$, RL algorithm $f_{RL}: \mathcal{M} \times i \times J \to \emph{path}$}
    \Output{SDR that includes initial and target nodes}
    \BlankLine
    \For{$n$ in range($N_w$)}{
        \For{each subnet $\mathcal{M}_S$ in $\mathcal{M}$}{
            \For{each node $j$ in $\mathcal{M}_S$}{
                \If{$reward_j > 0$}{
                \emph{$dynamic\_nodes \gets dynamic\_nodes \cup J$}}
            }
        \If{$dynamic\_nodes \neq \emptyset$}{
        \emph{$J \gets  \max(dynamic\_nodes) \cup J$}}
        }
    }
    \BlankLine
    \emph{$path \gets f_{RL}(\mathcal{M}, i, J)$}\;
    \BlankLine
    \Return paths
\end{algorithm}

\section{Experimental Design}

In the following subsections the network, state-action space, and RL algorithm implementation are described.

\subsection{Network Description}

The same network framework as in Cody et al \cite{cody2022discovering} is used for our experiment but with different configurations in defense mechanism. To simulate real-world network conditions, there are layers of defenses between the Internet and the innermost private network. Systems that require Internet to provide service (HTTP, email) are most vulnerable to attack;  and are typically in the Demilitarized Zone (DMZ) with limited access to private network resources. The private, internal networks are separated from the DMZ with firewalls that apply rules that only allow connections to specific internal servers/services. VPN services to the internal network is protected with VPN Management Firewalls that apply network rules allowing only authorized and authenticated user traffic to traverse internally over an encrypted connection.  Internal network subnets are separated based on access rules and allow traffic to egress to the Internet if it is authorized, as well as applying rules for network traversal between subnets. Finally, access to the innermost subnet is controlled by a firewall that allows only authorized traffic in or out, and only to specific hosts. 

These security controls were known and applied by the designers but were intentionally left unknown to the model to provide an accurate simulation of an attacker exploring an unknown network environment. 

\subsection{Environment Description}

The hosts are each represented as an 1D vector that encodes its status (compromised, reachable, discovered or not) and configurations (address, services, OS and processes). Next, our environment combines all vectors of hosts in the network as a entire state tensor. Our action space contains 6 primitive actions for scanning, exploiting and privilege escalation.

For these experiments, the initial host is set at (1, 0) while the terminal host is set as (3, 1), (8, 2) or (9, 2). Here, the host (x, y) refers to the host indexed by $y$ in subnet $x$. The initial host is compromised, reachable and discovered by default so that it allows the agent to perform further exploration, thus the simulation assumes the attacker already has gained this foothold in the network. The goal of SDR is to explore all the services of the target host. And a high positive reward (100) is assigned if the goal is reached. 

\subsection{RL Implementation}

The experiment is conducted on the single-agent A2C model and two variants of the double agent (DA) architecture\cite{nguyen2021proposal}. In the original double agent model, DA-A2C, both the structuring and the exploiting agents are trained using the A2C algorithm. To achieve better sample efficiency and training stability, we combine the PPO algorithm \cite{schulman2017proximal} with the DA architecture and build a DA-PPO model, where both RL agents are trained using PPO. 
We use Adam as the optimizer of our network. The A2C model is trained for 4000 episodes and the two double agent models are trained for 8000 episodes with a maximum of 3000 steps in each episode. The episode terminates either when the goal is reached or when the step limit is reached. Both the A2C model and the structuring agent of the DA models use deep neural networks (DNNs) with three fully connected layers of 64, 32 and 1 and the exploiting agent of the DA models use a DNN with two fully connected layers of size 10 and 1. All DNNs use tanh activation functions for non-output layers and softmax for the output layer.

\subsection{Sensitivity analysis}

We train the A2C, DA-A2C, DA-PPO to convergence. Our surveillance detection algorithm is run on four different scales values namely 1, 3, 5 and 11 for host (9,2) (8,2), and (3,1). For host (9,2) and (8,2) we run an extra scale of 15. The scale values were experimental in nature and each scale defines a certain drop in exploration of the network. When the scale value increases, the agent becomes more risk-averse and thus keeps on reducing the amount of exploration.

\begin{figure*}[htbp]
  \centering
  \includegraphics[width=0.8\linewidth]{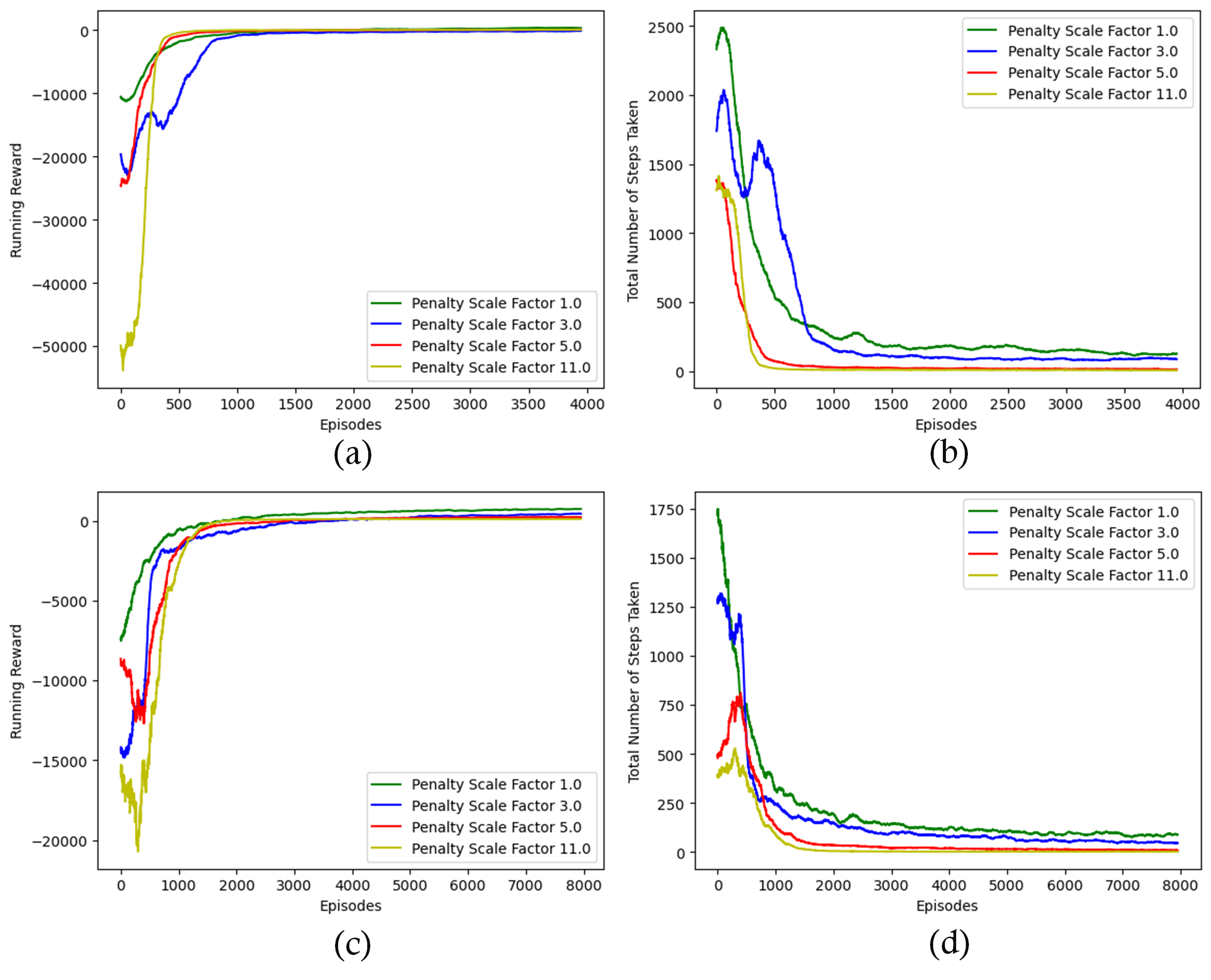}
  \caption{Training performance of DAA and A2C agent with different penalty scales: (a) episode-reward of A2C agent; (b) episode-step of A2C agent; (c) episode-reward of DAA agent and (d) episode-step of DAA agent.}\label{fig:A2C_vs_DAA}
\end{figure*}

\section{Results}

The model convergence is showcased by plotting the steps and reward versus episodes and results are shown in Figure \ref{fig:A2C_vs_DAA}. 

\begin{figure*}[t]
\begin{subfigure}{.5\textwidth}
  \centering
  \includegraphics[width=.95\linewidth]{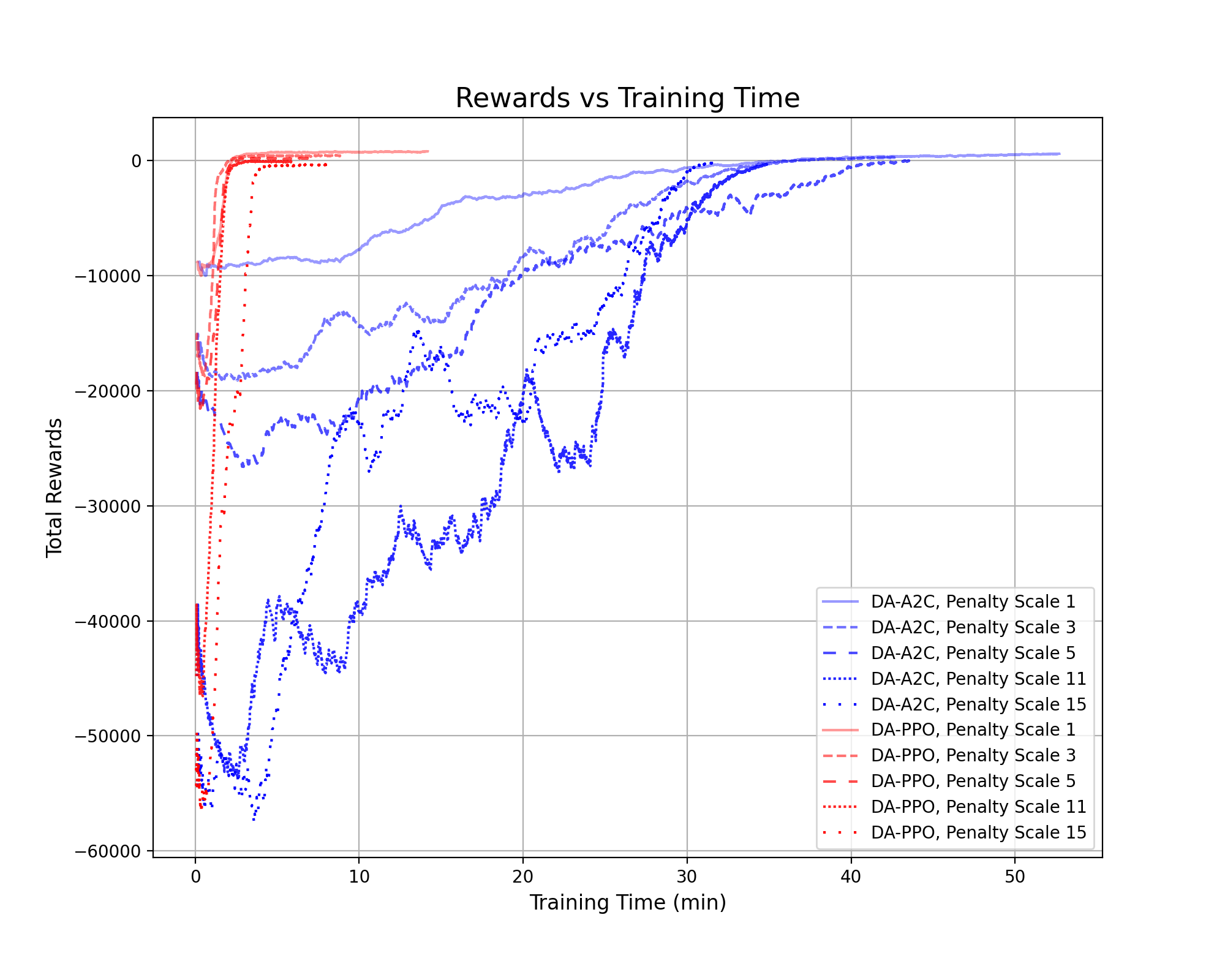}  
  \caption{}
  \label{fig:ppo_a2c_time}
\end{subfigure}
\begin{subfigure}{.5\textwidth}
  \centering
  \includegraphics[width=.95\linewidth]{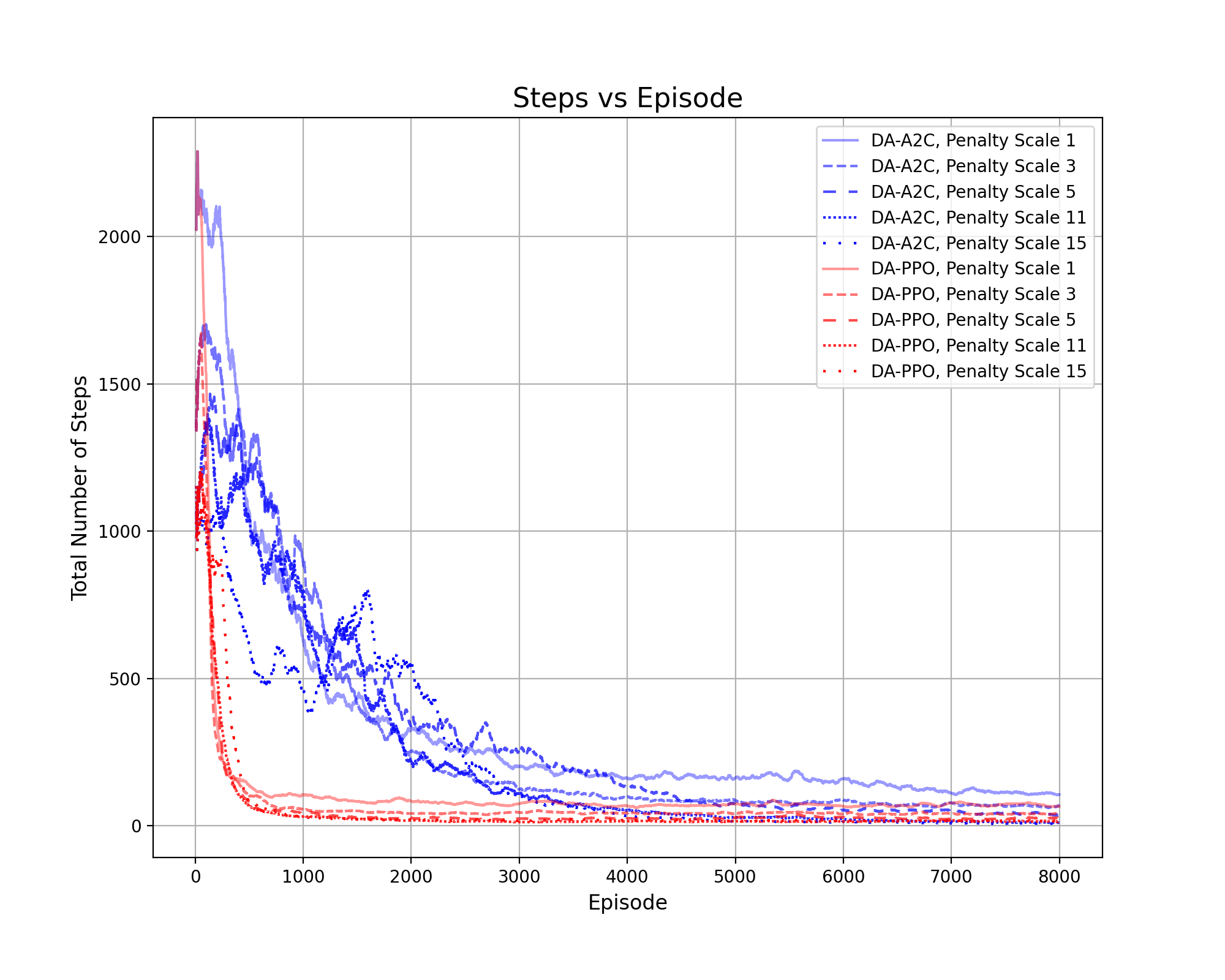}  
  \caption{}
  \label{fig:ppo_a2c_steps}
\end{subfigure}
\caption{Training performance of DA-A2C and DA-PPO with different penalty scales. The left plot shows the total rewards in an episode as training time increases. The right plot shows the total number of steps in each training episode.}
\label{fig:PPO_A2C}
\end{figure*}

As can be seen from Figure \ref{fig:PPO_A2C}, DA-PPO trains significantly faster than DA-A2C on all five penalty scales, both in terms of wall-time and number of episodes. Specifically, DA-PPO converges in less than ten minutes as compared to more than 30 minutes by DA-A2C (Figure \ref{fig:ppo_a2c_time}), and it requires much fewer episodes to learn an effective policy (Figure \ref{fig:ppo_a2c_steps}). 

\begin{table*}[t]
\centering
\begin{tabular}{ cccccc } 
\hline
Target Host & Scale Factor & Services & \makecell[c]{High\\Penalty Host}  & \makecell[c]{Medium\\Penalty Host} & \makecell[c]{Low\\Penalty Host} \\
\hline
\multirow{4}{4em}{(3, 1)} 
& 1 & 207 & 2 & 0 & 4 \\ 
& 3 & 166 & 1 & 0 & 4 \\ 
& 5 & 102 & 1 & 0 & 1 \\
& 11 & 72 & 1 & 0 & 1 \\
\hline
\multirow{4}{4em}{(8, 2)} 
& 1 & 207 & 2 & 0 & 4 \\ 
& 3 & 166 & 1 & 0 & 4 \\ 
& 5 & 134 & 1 & 1 & 2 \\
& 11 & 104 & 1 & 0 & 2 \\
\hline
\multirow{4}{4em}{(9, 2)} 
& 1 & 207 & 2 & 0 & 4 \\ 
& 3 & 166 & 1 & 0 & 4 \\ 
& 5 & 134 & 1 & 0 & 2 \\
& 11 & 104 & 1 & 0 & 1 \\ 
\hline
\end{tabular}
\caption{Table of the number of services and hosts explored with different target hosts and scale factors.}
\label{table:1}
\end{table*}

\begin{figure*}[htbp]
  \centering
  \includegraphics[width=0.8\linewidth]{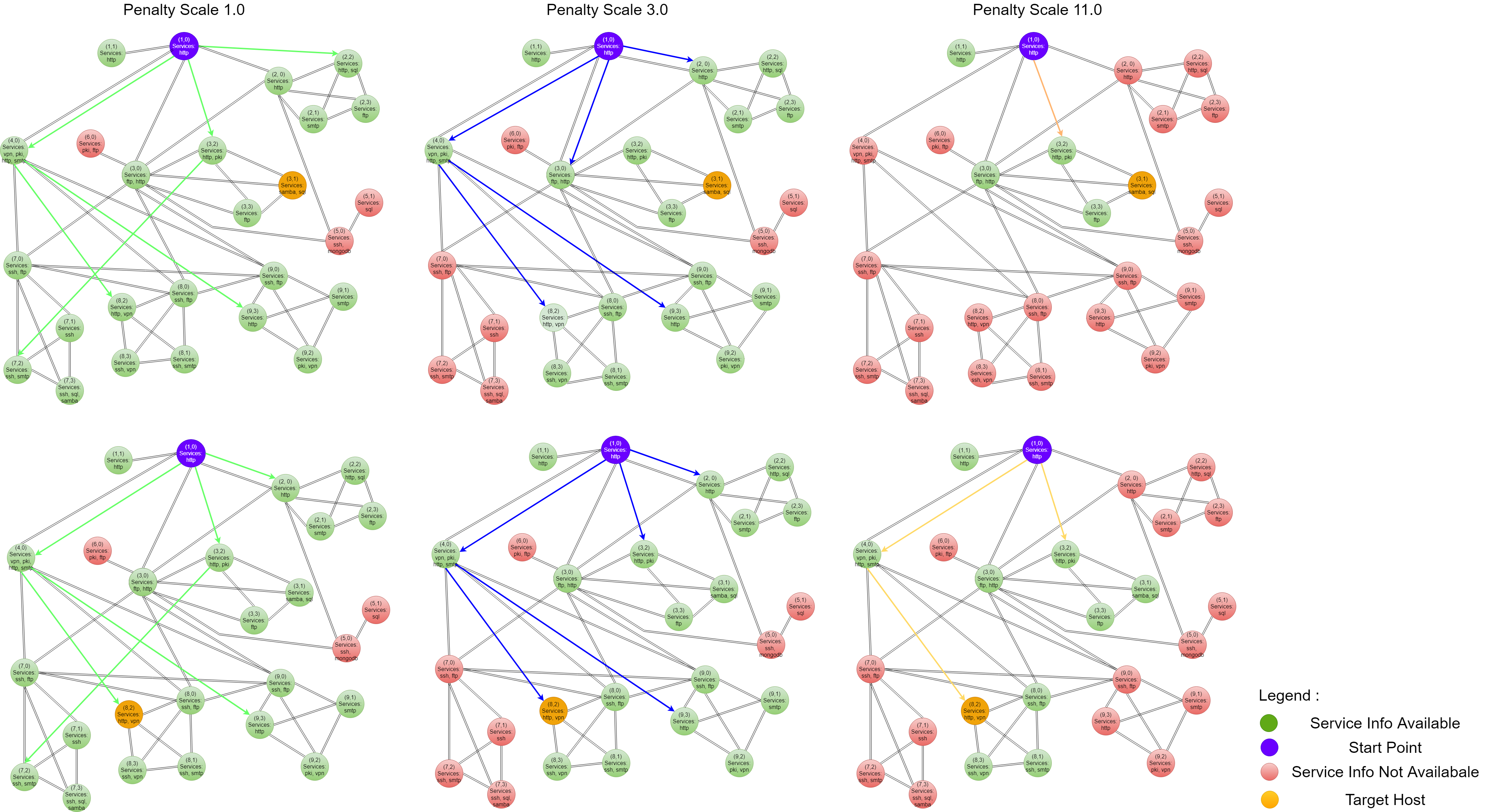}
  \caption{Network diagram showing the SDR to different target nodes for various penalty scale factors: host (3,1) for the first row and (8,2) for the second row. The nodes in green indicates the hosts whose service information is available while the information of nodes in red is not. The nodes in purple are set as initial nodes while the nodes in yellow are set as target nodes.}\label{fig:paths}
\end{figure*}

The DAA modeling highlights a pattern that showed the expected path being taken at all the penalty scales for all the three different target host, minimizing the number of defenses crossed. While the path converges by taking a relative higher number of episodes, it thus offers higher scalability while applying it to real world use cases with larger networks. Across the models, it is observed that as the penalty keeps on increasing the agent reduces the amount of exploration and only explored areas which are deemed to be safe. The simpler A2C model followed the expected path being taken in penalty scale of 1, 11 and 15 for three different target host. However, for scale of 3 and 5 in target host of (3,1), the A2C Model takes a potential imperfect path which expose to a higher risk; this suggests that even for a small test network of this size, a simpler agent scheme fails to find realistic paths.

\section{Discussion}

We can observe the paths taken by the agent to achieve the SDR goal when target hosts are (8,2) and (3,1) for penalty scales 1, 3 and 11 in Fig 3. In Table 1, we can observe that as we are adjusting the scale factor the number of services being explored drops off. Furthermore, the number of high penalty host controlled goes down from 2 to 1 as we scaled the penalty showcasing that the agent becomes more risk averse.

For penalty scale 1, the agent takes a path which according to the environment and penalty is acceptable as in its exploration of network it is taking control of hosts which are either low or medium risk except one high penalty host (i.e.,  (4,0)) which is a necessity as to reach the goal at least one high-risk host needs to be controlled. Therefore, the path taken by the agent for a low penalty of 1 is acceptable.

For penalty scale 3, when the target is (8,2) the agent takes an acceptable path according to the environment and penalty as it takes control of only low penalty host except one high penalty host (i.e., (4,0)) which is again a necessity to explore services on the target host. However, when the target host is (3,1), the agent takes a path which is less acceptable according to the environment and penalty as it takes control of two high penalty node (i.e., (4,0) and (3,0)) where only one was required to achieve the goal. The agent had an option to select (3,2) which is a lower penalty host than (3,0).

For penalty scale 11, the agent takes a path which is acceptable according the state and penalty of the environment and takes control of host which are low risk apart from only one high risk host which is unavoidable in achieving the goal.

Associating penalty scores (low to high) to the risk adversity of human actors demonstrates that the agent performs actions that are reasonably in line with expected human advisory behavior. 

Actors operating in a condensed timeline, commonly referred as “smash and grab” operations, or actors without sufficient experience, would be examples of an agent set with a Penalty Scale of 1.0. Nation-State actors, highly-competent APTs (Advanced Persistent Threats), and experienced actors with more time to observe and a higher cost of getting caught would be representative of an agent with a penalty scale of 11.0. 

When risk adversity is low, an actor is not worried about performing “noisy” scans. These scans will include enumeration of multiple network services or simultaneously traversing multiple network segments (with each new segment potentially having deeper layers of defense incorporated into it). As observed when the penalty score is set to one (1.0), the agent finds and explores paths that would involve a higher level of risk of getting caught and a lower sensitivity to negative consequences. These actions involve scanning or traversing paths with a reasonable presumption of greater security and more rigorously logged and monitored network devices, such as VPNs and Firewalls (FW) (High Risk Hosts) as well as enumerating networks not directly associated with the respective target. The risk-accepting paths also explored several networks unnecessarily and utilize multiple services along the way. Each service to be used and/or exploited along the way creates another risk and/or protection system that may be in place, thus increasing the log presence and actor footprint. (As stated above, it is presumed that firewalls are monitored at a higher rate, and that security services have the most inherited controls.) While exploitation of these devices yields great impact, the associated risk is also higher. This risky behavior is reflected by the agent as seen in Table 1 with the larger number of services and high penalty hosts being exploited, as well as represented by the paths chosen in Figure 3. 

When risk adversity is high, the cost of getting caught outweighs the reward of network discovery, so actors, and as observed, the agent, take the most direct path possible. Risk adverse actors naturally choose paths that avoid traversing multiple protocols/services and presumed increased security controls. This lowers the log footprint and network noise generated and reduces the chances of detection. The agent’s behavior of only taking control of one high risk host to reach the lower target reflects the lowest risk path that an advisory would naturally gravitate to.

\section{Conclusion}

In this work, we provide a RL realization of automating analysis methods for SDR. Our approach introduces a warm-up phase to perform pre-exploration emulating an experienced actors necessity to find the ``safe" areas of a network. The capability and efficiency of our methods are validated by simulations on a custom network configured with defense mechanisms (simulated cyber range). 

The model ran on a network created to simulate the realistic placement of layered network defenses with increasing security controls as an actor draws closer to the most sensitive information and services. As the model was faced with penalties that increased when transitioning into more secure zones, the effect of the penalty score became more demonstrative. This activity mirrors the expected risk sensitivity of an actor mirroring the risk profile of the model's penalty score. A more risk/penalty adverse model showed significantly more concise paths to reach the goal, while the less risk-adverse model explored around and left more of a footprint on the defense system’s logs. 

In future work, we expect to run this model on a live network utilizing both internal and external network data. Incorporating internal and external scan information allows for a model that most-closely mimics a malicious actor and defensive expectations for their actions. Using live data for a production network should also create greater granularity for the penalty scoring as vulnerability data for CVSS3 and weakness data for associated Common Weakness Enumeration (CWE) can be incorporated.

\bibliographystyle{IEEEtran}
\bibliography{ref}

\end{document}